\pdfoutput=1

\documentclass[11pt]{article}

\usepackage[]{acl}

\usepackage{times}
\usepackage{latexsym}
\usepackage{hyperref}

\usepackage[T1]{fontenc}

\usepackage[utf8]{inputenc}

\usepackage{microtype}

\title{L3Cube-MahaCorpus and MahaBERT: Marathi Monolingual Corpus, Marathi BERT Language Models, and Resources}

\author{Raviraj Joshi  \\
  Indian Institute of Technology Madras, Chennai \\
  L3Cube, Pune \\\
  \texttt{ravirajoshi@gmail.com}\\}

\begin{document}
\maketitle
\begin{abstract}
We present L3Cube-MahaCorpus a Marathi monolingual data set scraped from different internet sources. We expand the existing Marathi monolingual corpus with 24.8M sentences and 289M tokens. We further present, MahaBERT, MahaAlBERT, and MahaRoBerta all BERT-based masked language models, and MahaFT, the fast text word embeddings both trained on full Marathi corpus with 752M tokens. We show the effectiveness of these resources on downstream Marathi sentiment analysis, text classification, and named entity recognition (NER) tasks. We also release MahaGPT, a generative Marathi GPT model trained on Marathi corpus. Marathi is a popular language in India but still lacks these resources. This work is a step forward in building open resources for the Marathi language. The data and models are available at {\footnotesize https://github.com/l3cube-pune/MarathiNLP}.
\end{abstract}


\section{Introduction}

Pre-trained language models based on BERT have been widely used in NLP applications \cite{wolf2019huggingface,qiu2020pre}. These language models are fine-tuned on the target task and are reported to provide superior results. The target tasks include text classification, named entity recognition (NER), parts of speech (POS) tagging, dependency parsing, natural language inference (NLI), etc \cite{otter2020survey}. The BERT-based models can be trained using un-supervised large text corpus using masked language modeling objective and next sentence prediction tasks. 

The mono-lingual and multi-lingual masked language models have been very popular recently. The multi-lingual language models provide significant benefits for low resource languages by leveraging the learning from high resource text \cite{pires2019multilingual}. However, models trained on a single language are shown to perform better than multi-lingual models on target tasks in corresponding language \cite{straka2021robeczech}. Previous works have built BERT based language models in German, Vietnamese, Arabic, Dutch, French, Hindi, Bengali, etc \cite{scheible2020gottbert,nguyen2020phobert,le2019flaubert,delobelle2020robbert,abdul2020arbert,jain2020indic}. In this work, we focus on building monolingual corpus and BERT based language model in Marathi. Marathi is a low-resource Indian language and is native to the state of Maharashtra. 

Marathi is the third most popular language in India after Hindi and Bengali \cite{kulkarni2021experimental,joshi2019deep}. It is spoken by around 83 million people in India. Despite huge representation, in terms of speaking diaspora, the language resources have not received adequate attention for the Marathi language. The language resource in the simplest form is a monolingual corpus. However, even monolingual corpus for Indian languages is mostly biased towards Hindi. This can be seen from the fact that the recently released IndicNLP data set has 62.9M Hindi sentences and only 9.9M Marathi sentences \cite{kakwani2020inlpsuite}. There is a strong need to develop language resources for Marathi starting from building a monolingual corpus.

In this work, we add to the existing monolingual corpus by building L3Cube-MahaCorpus\footnote{https://github.com/l3cube-pune/MarathiNLP}. The data has been scraped from various internet sources. The corpus for Indian languages has mostly been exclusively dominated by news sources. We specifically consider this bias and also include sentences from non-news sources. L3Cube-MahaCorpus adds 24.8M sentences and 289M tokens (5.3 GB) to the existing Marathi monolingual datasets. After combining this with the existing Marathi corpus there is a total of 57.2M sentences and 752M tokens (13 GB). 

We further introduce MahaBERT\footnote{https://huggingface.co/l3cube-pune/marathi-bert}, MahaRoBERTa\footnote{https://huggingface.co/l3cube-pune/marathi-roberta}, MahaAlBERT\footnote{https://huggingface.co/l3cube-pune/marathi-albert} \footnote{https://huggingface.co/l3cube-pune/marathi-albert-v2}, and MahaGPT\footnote{https://huggingface.co/l3cube-pune/marathi-gpt} all Transformer BERT based Marathi language models trained on the full Marathi monolingual corpus. The BERT models are trained using masked language modeling objectives. These models are further evaluated on downstream tasks of text classification and named entity recognition (NER) in Marathi.
We also release MahaFT, the fast text word embedding trained on the full Marathi Corpus.
The dataset and resources are publicly shared to facilitate further research in Marathi NLP. 
The main contributions of this work are:
\begin{itemize}
    \item We present L3Cube-MahaCorpus, a Marathi monolingual corpus with 24.8M sentences and 289M tokens.
    \item We introduce MahaBERT,  MahaAlBERT, and MahaRoBERTa, the BERT variations trained on a full corpus with 752M tokens. We also release MahaGPT, a Marathi generative pre-trained transformer model trained on the full corpus.
    \item Finally, we release MahaFT, Marathi fast text embeddings trained on the full corpus. 
\end{itemize}

\section{Related Work}

In this section, we review different unsupervised and supervised data sets in the Marathi language. A summary of publicly available Marathi monolingual corpus and classification data sets is provided in \newcite{kulkarni2021experimental}. The main sources include Wikipedia text, CC-100 Dataset \cite{wenzek2019ccnet}, OSCAR Corpus \cite{suarez2019asynchronous}, and IndicNLP Corpus \cite{kakwani2020inlpsuite}. The wiki dataset consists of 85k cleaned Marathi articles. The other sources are multi-lingual datasets with Marathi as one of the languages. The CC-100 monolingual data set consists of around 50 million tokens for the Marathi language. The OSCAR corpus consists of around 82 million tokens in Marathi. The IndicNLP is perhaps the largest non-wiki source and consists of 142 million tokens.   

There are limited resources for supervised tasks in Marathi. The text classification data set includes IndicNLP News Article Dataset \cite{kakwani2020inlpsuite}, iNLTK Headline Dataset \cite{arora2020inltk}, L3CubeMahaSent \cite{kulkarni2021l3cubemahasent}. The IndicNLP News Article Dataset is a news article classification dataset in Marathi consisting of 4779 records. The iNLTK Headline Dataset categorizes news headlines and consists of 12092 records. The L3CubeMahaSent is a sentiment classification dataset in Marathi and consists of 16000 records.
Another data set for Marathi NER was introduced in \cite{murthy2018judicious}. It consists of 5591 sentences and 3 named entities as target labels.
Moreover, some hate speech detection datasets have also been released in Marathi \cite{gaikwad2021cross,mandl2021overview,pawar2019multilingual}.

In this work, we have utilized all publicly available Marathi monolingual corpus along with the L3Cube-MahaCorpus to train the language models and word embedding. These models are evaluated on the three classification tasks and a NER task.

\section{Curation of Dataset}

The L3Cube-MahaCorpus is collected from news and non-news sources. The major chunk of the data is scraped from the Maharashtra Times website \footnote{https://maharashtratimes.com/}. The non-news sources were taken from a collection website \footnote{http://www.netshika.com/sangrah.html}. The data set was scraped using the BeautifulSoup library along with the use of Selenium for dynamic pages. The final data set was shuffled and de-duplicated. The de-duplication was also performed with the existing monolingual data set. The L3Cube-MahaCorpus adds 17.6 M sentences (212 M tokens) from the news sources and 7.2 M sentences (76.4 M tokens) from the non-news sources. These are made available separately as well. Overall it adds 24.8 M sentences and 289 M tokens. When combined with the existing monolingual dataset, we now have 57.2 M sentences and 752 M tokens in the Marathi language. These statistics are also described in Table \ref{stat-tab}.

\begin{table}
\begin{center}
\begin{tabular}{{p{3.5cm}p{1cm}p{1.5cm}}}
\hline
\textbf{Dataset}&\textbf{\#tokens}&\textbf{\#sentences} \\
\hline
L3Cube-MahaCorpus (News)& 212 & 17.6  \\
L3Cube-MahaCorpus (Non-news) & 76.4 & 7.2 \\
L3Cube-MahaCorpus & 289 & 24.8 \\
Full Marathi Corpus & 752 & 57.2  \\
\hline
\end{tabular}
\caption{\label{stat-tab} Dataset Statistics (in millions). }
\end{center}
\end{table}

\section{Pre-trained Resources}
The full Marathi monolingual corpus is used to train Transformer based masked language models and FastText word embeddings. 
\subsection{Transformer Models}
The BERT represents a deep bi-directional Transformer based model trained using a large unlabelled corpus. These pre-trained models have been shown to produce state-of-the-art results on a variety of downstream tasks. There are different variations of BERT models like AlBERT and RoBERTa which are also considered in this work. From the multilingual perspective, there are three main models which can also be used with the Marathi language. These include multilingual-BERT \cite{devlin2018bert},  XLM-R based on RoBERTa \cite{conneau2019unsupervised}, and IndicBERT \cite{kakwani2020inlpsuite} based on AlBERT. These three models are fine-tuned on monolingual Marathi corpus and released as a part of this work. All the models are trained for 2 epochs with standard hyper-parameters and masked language modeling objective only. The learning rate used is 2e-5 with a batch size of 64. 
\begin{itemize}
    \item mBERT\footnote{\scriptsize https://huggingface.co/bert-base-multilingual-cased}: It is a BERT-base vanilla model pre-trained on 104 languages using masked language modeling (MLM) and next sentence prediction (NSP) objective. The Marathi was one of the languages used in pre-training.
    \item XLM-RoBERTa\footnote{\scriptsize	 https://huggingface.co/docs/transformers/model\_doc/xlmroberta}: It is a RoBERTa based model pre-trained on 100 languages using MLM objective. The model is shown to outperform mBERT on different tasks. Even this model contains Marathi as one of the pre-training languages. The RoBERTa mainly modifies the hyper-parameters used in the original BERT and gets rid of the NSP task \cite{liu2019roberta}.
    \item IndicBERT\footnote{\scriptsize https://huggingface.co/ai4bharat/indic-bert}: It is a multi-lingual AlBERT model exclusively pre-trained on 12 Indian languages. The AlBERT is a lite version of the BERT model \cite{lan2019albert}. It uses parameter reduction techniques like repeated layers to reduce the memory footprint. The model has been shown to work well on most of the Indic NLP tasks \cite{joshi2021evaluation,kulkarni2021l3cubemahasent,velankar2021hate,nayak2021contextual}.
\end{itemize}
\subsection{FastText Word Embeddings}
Pre-trained word embeddings are commonly used to initialize the embedding layer of the neural networks. These distributed representations are trained on large unlabeled corpus and are useful for many downstream tasks. The FastText word embeddings are popular for morphologically rich languages \cite{bojanowski2017enriching}. It represents the word as a bag of character n-grams thus avoiding any out of vocabulary word. We train the FastText model on the Marathi monolingual corpus using standard hyper-parameters. A skip-gram model is trained with a window size of 5, 10 negative samples per instance, and 10 epochs. 
\subsection{Marathi GPT}
GPT2 is a generative transformer model trained using causal language modeling (CLM) objective \cite{radford2019language}. It is also a class of self-supervised models trained to predict the next work on the unsupervised data. We train a standard GPT2 model with 12 layers and 768 internal dimension on Marathi Corpus for 5 epochs with a learning rate of 2e-5. We use a custom BPE-based tokenizer with a vocab size of 50257.

    \begin{table*}
\begin{center}
\begin{tabular}{{p{3cm}p{3cm}p{3cm}p{2cm}p{2cm}}}
\hline \textbf{Model} & \textbf{L3CubeMahaSent} & \textbf{News Articles} & \textbf{News Headlines} & \textbf{Marathi NER} \\
\hline
mBERT & 80.4 & 97.6 & 90.6 & 58.35\\
indicBERT &  83.3 & 98.7 & 93.7 & 60.79\\
XLM-R & 82.0 & 98.5 & 92.5 & 62.32\\ \hline
MahaBERT &  82.8 & 98.7 & 94.4 & 62.57\\ 
MahaAlBERT & \textbf{83.7} & \textbf{99.1} & \textbf{94.7} & 60.00\\ 
MahaRoBERTa & 83.4 & 98.5 & 94.2 & \textbf{64.34}\\ \hline
FB-FT + KNN & 73.6 & 99.1 & 88.8 & -\\ 
INLP-FT + KNN & 74.9 & 98.9 & 90.7 & -\\
MahaFT + KNN & 75.1 & 98.9 & 91.2 & -\\ \hline

\hline
\end{tabular}
\caption{\label{result-tab} The results for different models on classification and NER tasks. The numbers for classification task L3CubeMahaSent, News Articles, and News Headlines represent the classification accuracy. The numbers for the Marathi NER task represent the macro-f1 score. The FB-FT is Marathi fast text embeddings trained on Wiki and Common Crawl Corpus released by Facebook used along with KNN(k=4). The INLP-FT represents the Marathi fast text embeddings released by IndicNLP Suite. The MahaFT are Marathi fast text embeddings released as a part of this work.}
\end{center}
\end{table*}

\section{Down Stream Tasks}
\begin{itemize}
    \item \textbf{IndicNLP News Article Classification}: The task consists of Marathi news articles classified as sports, entertainment, and lifestyle. There are 3823 train, 479 test, and 477 validation examples.
    \item \textbf{iNLTK Headline Classification}: In this classification task the Marathi news headlines are categorized as entertainment, sports, and state. The dataset consists of 9672 train, 1210 test, and 1210 validation examples.
    \item \textbf{L3CubeMahaSent Sentiment Analysis}: The sentiment analysis task consists of Marathi tweets categorized as positive, negative, and neutral. The dataset consists of 12114 train, 2250 test, and 1500 validation examples.
    \item \textbf{Marathi Named Entity Recognition}: This is a Marathi entity recognition task where each token in the sentence is categorized as Location, Person, and Organization. The dataset consists of 3588 train, 1533 test, and 470 validation examples.

\end{itemize}

\subsection{Results}
The L3Cube-MahaCorpus along with other publicly available Marathi corpus is used to train three variations of BERT using MLM objective. These variations are based on base-BERT, AlBERT, and RoBERTa architecture and are termed as MahaBERT, MahaAlBERT, and MahaRoBERTa respectively. The multilingual versions of these architectures mBERT, indicBERT based on AlBERT, and XLM-R based on RoBERTa are also used for baseline comparison. The multilingual versions are fine-tuned on the Marathi corpus to get the Marathi BERT models. Similar hyper-parameters are used for MLM pre-training of all these models. 
The results are described in Table \ref{result-tab}. Note that the results for base models may be slightly different than ones reported in the original work as they were re-computed using a common setup and hyperparameters. These models are evaluated on three classification datasets and one named entity recognition dataset. For the classification task, the pre-trained models are further fine-tuned by the addition of a dense layer on top of [CLS] token embedding. The NER task is formulated as a token classification task and all token embeddings are passed through the dense layer for classification.  Overall the monolingual versions of models perform better than the multi-lingual versions. 

The fast text word embeddings trained on full Marathi corpus termed as MahaFT are evaluated on the classification datasets. In this setup, word embeddings are averaged to get the sentence representation. A KNN classifier with k=4 is used for the classification of the averaged fast text embedding. These Marathi word embeddings are compared against two other publicly available variations. The FB-FT represents Marathi fast text embeddings trained on Wiki and Common Crawl Corpus released by Facebook. The INLP-FT was released as part of IndicNLP suite. The MahaFT performs competitively with other word embeddings. Overall we show the resources released as a part of this work either perform competitively with or better than the currently available alternatives for the Marathi language.

\section{Conclusion}

In this paper, we have presented L3Cube-MahaCorpus, MahaBERT, and MahaFT. The MahaCorpus, is a Marathi monolingual corpus and is a significant addition to the existing monolingual corpus. The Marathi BERT is trained in three different flavors namely MahaBERT, MahaRoBERTa, and MahaAlBERT. The MahaFT is the Marathi fast text word embeddings. These resources are exclusively trained on Marathi monolingual corpus. The models are evaluated on downstream Marathi classification and NER tasks. The models are shown to work better than their multi-lingual counterparts. 

\section*{Acknowledgments}

Multiple L3Cube Pune, student groups have contributed to this work. We would like to thank Atharva Kulkarni, Meet Mandhane, Manali Likhitkar, and Gayatri Kshirsagar for their contribution. We also thank groups Algorithm\_Unlock and Bits\_To\_Bytes for their support. 

\bibliography{main}

\begin{thebibliography}{31}
\expandafter\ifx\csname natexlab\endcsname\relax\def\natexlab#1{#1}\fi

\bibitem[{Abdul-Mageed et~al.(2020)Abdul-Mageed, Elmadany, and
  Nagoudi}]{abdul2020arbert}
Muhammad Abdul-Mageed, AbdelRahim Elmadany, and El~Moatez~Billah Nagoudi. 2020.
\newblock Arbert \& marbert: deep bidirectional transformers for arabic.
\newblock \emph{arXiv preprint arXiv:2101.01785}.

\bibitem[{Arora(2020)}]{arora2020inltk}
Gaurav Arora. 2020.
\newblock inltk: Natural language toolkit for indic languages.
\newblock \emph{arXiv preprint arXiv:2009.12534}.

\bibitem[{Bojanowski et~al.(2017)Bojanowski, Grave, Joulin, and
  Mikolov}]{bojanowski2017enriching}
Piotr Bojanowski, Edouard Grave, Armand Joulin, and Tomas Mikolov. 2017.
\newblock Enriching word vectors with subword information.
\newblock \emph{Transactions of the Association for Computational Linguistics},
  5:135--146.

\bibitem[{Conneau et~al.(2019)Conneau, Khandelwal, Goyal, Chaudhary, Wenzek,
  Guzm{\'a}n, Grave, Ott, Zettlemoyer, and Stoyanov}]{conneau2019unsupervised}
Alexis Conneau, Kartikay Khandelwal, Naman Goyal, Vishrav Chaudhary, Guillaume
  Wenzek, Francisco Guzm{\'a}n, Edouard Grave, Myle Ott, Luke Zettlemoyer, and
  Veselin Stoyanov. 2019.
\newblock Unsupervised cross-lingual representation learning at scale.
\newblock \emph{arXiv preprint arXiv:1911.02116}.

\bibitem[{Delobelle et~al.(2020)Delobelle, Winters, and
  Berendt}]{delobelle2020robbert}
Pieter Delobelle, Thomas Winters, and Bettina Berendt. 2020.
\newblock Robbert: a dutch roberta-based language model.
\newblock \emph{arXiv preprint arXiv:2001.06286}.

\bibitem[{Devlin et~al.(2019)Devlin, Chang, Lee, and
  Toutanova}]{devlin2018bert}
Jacob Devlin, Ming-Wei Chang, Kenton Lee, and Kristina Toutanova. 2019.
\newblock Bert: Pre-training of deep bidirectional transformers for language
  understanding.
\newblock In \emph{Proceedings of the 2019 Conference of the North American
  Chapter of the Association for Computational Linguistics: Human Language
  Technologies, Volume 1 (Long and Short Papers)}, pages 4171--4186.

\bibitem[{Gaikwad et~al.(2021)Gaikwad, Ranasinghe, Zampieri, and
  Homan}]{gaikwad2021cross}
Saurabh~Sampatrao Gaikwad, Tharindu Ranasinghe, Marcos Zampieri, and
  Christopher Homan. 2021.
\newblock Cross-lingual offensive language identification for low resource
  languages: The case of marathi.
\newblock In \emph{Proceedings of the International Conference on Recent
  Advances in Natural Language Processing (RANLP 2021)}, pages 437--443.

\bibitem[{Jain et~al.(2020)Jain, Deshpande, Shridhar, Laumann, and
  Dash}]{jain2020indic}
Kushal Jain, Adwait Deshpande, Kumar Shridhar, Felix Laumann, and Ayushman
  Dash. 2020.
\newblock Indic-transformers: An analysis of transformer language models for
  indian languages.
\newblock \emph{arXiv preprint arXiv:2011.02323}.

\bibitem[{Joshi et~al.(2019)Joshi, Goel, and Joshi}]{joshi2019deep}
Ramchandra Joshi, Purvi Goel, and Raviraj Joshi. 2019.
\newblock Deep learning for hindi text classification: A comparison.
\newblock In \emph{International Conference on Intelligent Human Computer
  Interaction}, pages 94--101. Springer.

\bibitem[{Joshi et~al.(2021)Joshi, Karnavat, Jirapure, and
  Joshi}]{joshi2021evaluation}
Ramchandra Joshi, Rushabh Karnavat, Kaustubh Jirapure, and Ravirai Joshi. 2021.
\newblock Evaluation of deep learning models for hostility detection in hindi
  text.
\newblock In \emph{2021 6th International Conference for Convergence in
  Technology (I2CT)}, pages 1--5. IEEE.

\bibitem[{Kakwani et~al.(2020)Kakwani, Kunchukuttan, Golla, Gokul,
  Bhattacharyya, Khapra, and Kumar}]{kakwani2020inlpsuite}
Divyanshu Kakwani, Anoop Kunchukuttan, Satish Golla, NC~Gokul, Avik
  Bhattacharyya, Mitesh~M Khapra, and Pratyush Kumar. 2020.
\newblock inlpsuite: Monolingual corpora, evaluation benchmarks and pre-trained
  multilingual language models for indian languages.
\newblock In \emph{Proceedings of the 2020 Conference on Empirical Methods in
  Natural Language Processing: Findings}, pages 4948--4961.

\bibitem[{Kulkarni et~al.(2021{\natexlab{a}})Kulkarni, Mandhane, Likhitkar,
  Kshirsagar, Jagdale, and Joshi}]{kulkarni2021experimental}
Atharva Kulkarni, Meet Mandhane, Manali Likhitkar, Gayatri Kshirsagar,
  Jayashree Jagdale, and Raviraj Joshi. 2021{\natexlab{a}}.
\newblock Experimental evaluation of deep learning models for marathi text
  classification.
\newblock \emph{arXiv preprint arXiv:2101.04899}.

\bibitem[{Kulkarni et~al.(2021{\natexlab{b}})Kulkarni, Mandhane, Likhitkar,
  Kshirsagar, and Joshi}]{kulkarni2021l3cubemahasent}
Atharva Kulkarni, Meet Mandhane, Manali Likhitkar, Gayatri Kshirsagar, and
  Raviraj Joshi. 2021{\natexlab{b}}.
\newblock L3cubemahasent: A marathi tweet-based sentiment analysis dataset.
\newblock In \emph{Proceedings of the Eleventh Workshop on Computational
  Approaches to Subjectivity, Sentiment and Social Media Analysis}, pages
  213--220.

\bibitem[{Lan et~al.(2019)Lan, Chen, Goodman, Gimpel, Sharma, and
  Soricut}]{lan2019albert}
Zhenzhong Lan, Mingda Chen, Sebastian Goodman, Kevin Gimpel, Piyush Sharma, and
  Radu Soricut. 2019.
\newblock Albert: A lite bert for self-supervised learning of language
  representations.
\newblock \emph{arXiv preprint arXiv:1909.11942}.

\bibitem[{Le et~al.(2019)Le, Vial, Frej, Segonne, Coavoux, Lecouteux, Allauzen,
  Crabb{\'e}, Besacier, and Schwab}]{le2019flaubert}
Hang Le, Lo{\"\i}c Vial, Jibril Frej, Vincent Segonne, Maximin Coavoux,
  Benjamin Lecouteux, Alexandre Allauzen, Beno{\^\i}t Crabb{\'e}, Laurent
  Besacier, and Didier Schwab. 2019.
\newblock Flaubert: Unsupervised language model pre-training for french.
\newblock \emph{arXiv preprint arXiv:1912.05372}.

\bibitem[{Liu et~al.(2019)Liu, Ott, Goyal, Du, Joshi, Chen, Levy, Lewis,
  Zettlemoyer, and Stoyanov}]{liu2019roberta}
Yinhan Liu, Myle Ott, Naman Goyal, Jingfei Du, Mandar Joshi, Danqi Chen, Omer
  Levy, Mike Lewis, Luke Zettlemoyer, and Veselin Stoyanov. 2019.
\newblock Roberta: A robustly optimized bert pretraining approach.
\newblock \emph{arXiv preprint arXiv:1907.11692}.

\bibitem[{Mandl et~al.(2021)Mandl, Modha, Shahi, Madhu, Satapara, Majumder,
  Schaefer, Ranasinghe, Zampieri, Nandini et~al.}]{mandl2021overview}
Thomas Mandl, Sandip Modha, Gautam~Kishore Shahi, Hiren Madhu, Shrey Satapara,
  Prasenjit Majumder, Johannes Schaefer, Tharindu Ranasinghe, Marcos Zampieri,
  Durgesh Nandini, et~al. 2021.
\newblock Overview of the hasoc subtrack at fire 2021: Hate speech and
  offensive content identification in english and indo-aryan languages.
\newblock \emph{arXiv preprint arXiv:2112.09301}.

\bibitem[{Murthy et~al.(2018)Murthy, Kunchukuttan, and
  Bhattacharyya}]{murthy2018judicious}
Rudra Murthy, Anoop Kunchukuttan, and Pushpak Bhattacharyya. 2018.
\newblock Judicious selection of training data in assisting language for
  multilingual neural ner.
\newblock In \emph{Proceedings of the 56th Annual Meeting of the Association
  for Computational Linguistics (Volume 2: Short Papers)}, pages 401--406.

\bibitem[{Nayak and Joshi(2021)}]{nayak2021contextual}
Ravindra Nayak and Raviraj Joshi. 2021.
\newblock Contextual hate speech detection in code mixed text using transformer
  based approaches.
\newblock \emph{arXiv preprint arXiv:2110.09338}.

\bibitem[{Nguyen and Nguyen(2020)}]{nguyen2020phobert}
Dat~Quoc Nguyen and Anh~Tuan Nguyen. 2020.
\newblock Phobert: Pre-trained language models for vietnamese.
\newblock \emph{arXiv preprint arXiv:2003.00744}.

\bibitem[{Otter et~al.(2020)Otter, Medina, and Kalita}]{otter2020survey}
Daniel~W Otter, Julian~R Medina, and Jugal~K Kalita. 2020.
\newblock A survey of the usages of deep learning for natural language
  processing.
\newblock \emph{IEEE Transactions on Neural Networks and Learning Systems},
  32(2):604--624.

\bibitem[{Pawar and Raje(2019)}]{pawar2019multilingual}
Rohit Pawar and Rajeev~R Raje. 2019.
\newblock Multilingual cyberbullying detection system.
\newblock In \emph{2019 IEEE International Conference on Electro Information
  Technology (EIT)}, pages 040--044. IEEE.

\bibitem[{Pires et~al.(2019)Pires, Schlinger, and
  Garrette}]{pires2019multilingual}
Telmo Pires, Eva Schlinger, and Dan Garrette. 2019.
\newblock How multilingual is multilingual bert?
\newblock \emph{arXiv preprint arXiv:1906.01502}.

\bibitem[{Qiu et~al.(2020)Qiu, Sun, Xu, Shao, Dai, and Huang}]{qiu2020pre}
Xipeng Qiu, Tianxiang Sun, Yige Xu, Yunfan Shao, Ning Dai, and Xuanjing Huang.
  2020.
\newblock Pre-trained models for natural language processing: A survey.
\newblock \emph{Science China Technological Sciences}, pages 1--26.

\bibitem[{Radford et~al.(2019)Radford, Wu, Child, Luan, Amodei, Sutskever
  et~al.}]{radford2019language}
Alec Radford, Jeffrey Wu, Rewon Child, David Luan, Dario Amodei, Ilya
  Sutskever, et~al. 2019.
\newblock Language models are unsupervised multitask learners.
\newblock \emph{OpenAI blog}, 1(8):9.

\bibitem[{Scheible et~al.(2020)Scheible, Thomczyk, Tippmann, Jaravine, and
  Boeker}]{scheible2020gottbert}
Raphael Scheible, Fabian Thomczyk, Patric Tippmann, Victor Jaravine, and Martin
  Boeker. 2020.
\newblock Gottbert: a pure german language model.
\newblock \emph{arXiv preprint arXiv:2012.02110}.

\bibitem[{Straka et~al.(2021)Straka, N{\'a}plava, Strakov{\'a}, and
  Samuel}]{straka2021robeczech}
Milan Straka, Jakub N{\'a}plava, Jana Strakov{\'a}, and David Samuel. 2021.
\newblock Robeczech: Czech roberta, a monolingual contextualized language
  representation model.
\newblock \emph{arXiv preprint arXiv:2105.11314}.

\bibitem[{Su{\'a}rez et~al.(2019)Su{\'a}rez, Sagot, and
  Romary}]{suarez2019asynchronous}
Pedro Javier~Ortiz Su{\'a}rez, Beno{\^\i}t Sagot, and Laurent Romary. 2019.
\newblock Asynchronous pipeline for processing huge corpora on medium to low
  resource infrastructures.
\newblock In \emph{7th Workshop on the Challenges in the Management of Large
  Corpora (CMLC-7)}. Leibniz-Institut f{\"u}r Deutsche Sprache.

\bibitem[{Velankar et~al.(2021)Velankar, Patil, Gore, Salunke, and
  Joshi}]{velankar2021hate}
Abhishek Velankar, Hrushikesh Patil, Amol Gore, Shubham Salunke, and Raviraj
  Joshi. 2021.
\newblock Hate and offensive speech detection in hindi and marathi.
\newblock \emph{arXiv preprint arXiv:2110.12200}.

\bibitem[{Wenzek et~al.(2019)Wenzek, Lachaux, Conneau, Chaudhary, Guzm{\'a}n,
  Joulin, and Grave}]{wenzek2019ccnet}
Guillaume Wenzek, Marie-Anne Lachaux, Alexis Conneau, Vishrav Chaudhary,
  Francisco Guzm{\'a}n, Armand Joulin, and Edouard Grave. 2019.
\newblock Ccnet: Extracting high quality monolingual datasets from web crawl
  data.
\newblock \emph{arXiv preprint arXiv:1911.00359}.

\bibitem[{Wolf et~al.(2019)Wolf, Debut, Sanh, Chaumond, Delangue, Moi, Cistac,
  Rault, Louf, Funtowicz et~al.}]{wolf2019huggingface}
Thomas Wolf, Lysandre Debut, Victor Sanh, Julien Chaumond, Clement Delangue,
  Anthony Moi, Pierric Cistac, Tim Rault, R{\'e}mi Louf, Morgan Funtowicz,
  et~al. 2019.
\newblock Huggingface's transformers: State-of-the-art natural language
  processing.
\newblock \emph{arXiv preprint arXiv:1910.03771}.

\end{thebibliography}
\bibliographystyle{acl_natbib}




\end{document}